\documentclass[letterpaper, 10pt, conference]{IEEEtran}      
\IEEEoverridecommandlockouts

\usepackage[utf8]{inputenc}
\usepackage[english]{babel}
\usepackage[T1]{fontenc}
\usepackage{amssymb,amsfonts}
\usepackage[cmex10]{amsmath}
\usepackage{dsfont}
\usepackage{algorithm}
\usepackage{algpseudocode}
\usepackage{multirow}
\usepackage{graphicx}
\usepackage{textcomp}
\usepackage{xcolor}
\usepackage[binary-units=true]{siunitx}
\usepackage{tikz}
\usepackage[caption=false,font=footnotesize]{subfig}
\usepackage{hyperref}
\usepackage[capitalize]{cleveref}
\def\BibTeX{{\rm B\kern-.05em{\sc i\kern-.025em b}\kern-.08em
    T\kern-.1667em\lower.7ex\hbox{E}\kern-.125emX}}

\newcommand{\disable}[1]{}
\newcommand{\Tmax}{\ensuremath{\text{C}\textsubscript{max}}}
\newcommand{\Dmax}{\ensuremath{\text{D}\textsubscript{max}}}
\newcommand{\Cmax}{\ensuremath{\text{C}\textsubscript{max}}}
\newcommand{\Vmax}{\ensuremath{\text{v}\textsubscript{max}}}
\newcommand{\Vinsp}{\ensuremath{\text{v}\textsubscript{insp}}}

\newcommand{\boundellipse}[3]
{(#1) ellipse (#2 and #3)
}

\usepackage{scalerel}
\usetikzlibrary{svg.path}
\definecolor{orcidlogocol}{HTML}{A6CE39}
\tikzset{
orcidlogo/.pic={
  \fill[orcidlogocol] svg{M256,128c0,70.7-57.3,128-128,128C57.3,256,0,198.7,0,128C0,57.3,57.3,0,128,0C198.7,0,256,57.3,256,128z};
  \fill[white] svg{M86.3,186.2H70.9V79.1h15.4v48.4V186.2z}
               svg{M108.9,79.1h41.6c39.6,0,57,28.3,57,53.6c0,27.5-21.5,53.6-56.8,53.6h-41.8V79.1z M124.3,172.4h24.5c34.9,0,42.9-26.5,42.9-39.7c0-21.5-13.7-39.7-43.7-39.7h-23.7V172.4z}
               svg{M88.7,56.8c0,5.5-4.5,10.1-10.1,10.1c-5.6,0-10.1-4.6-10.1-10.1c0-5.6,4.5-10.1,10.1-10.1C84.2,46.7,88.7,51.3,88.7,56.8z};
}
}
\newcommand\orcidicon[1]{\href{https://orcid.org/#1}{\mbox{\scalerel*{
\begin{tikzpicture}[yscale=-1,transform shape]
\pic{orcidlogo};
\end{tikzpicture}
}{|}}}}

\begin{document}
\bstctlcite{mstsp-bib}

\title{Vehicle Fault-Tolerant Robust Power Transmission Line Inspection Planning}

\author{František Nekovář$^{\orcidicon{0000-0002-1975-078X}}$, \and Jan Faigl$^{\orcidicon{0000-0002-6193-0792}}$, \and Martin Saska$^{\orcidicon{0000-0001-7106-3816}}$
\thanks{
  This work was supported by the Czech Science Foundation (GAČR) under research projects No. 22-05762S and No. 22-24425S, and by the European Union’s Horizon 2020 research and innovation programme AERIAL-CORE under grant agreement no. 871479.}
\thanks{Authors are with the Czech Technical University, Faculty of Electrical Engineering, Technicka 2, 166 27, Prague, Czech Republic, email: {\tt\small \{nekovfra|faiglj|saskam1\}@fel.cvut.cz}.}
}
\maketitle

\begin{abstract}
   This paper concerns fault-tolerant power transmission line inspection planning as a generalization of the multiple traveling salesmen problem.  
The addressed inspection planning problem is formulated as a single-depot multiple-vehicle scenario, where the inspection vehicles are constrained by the battery budget limiting their inspection time.
The inspection vehicle is assumed to be an autonomous multi-copter with a wide range of possible flight speeds influencing battery consumption.
The inspection plan is represented by multiple routes for vehicles providing full coverage over inspection target power lines.
On an inspection vehicle mission interruption, which might happen at any time during the execution of the inspection plan, the inspection is re-planned using the remaining vehicles and their remaining battery budgets.
Robustness is introduced by choosing a suitable cost function for the initial plan that maximizes the time window for successful re-planning.
It enables the remaining vehicles to successfully finish all the inspection targets using their respective remaining battery budgets.
A combinatorial metaheuristic algorithm with various cost functions is used for planning and fast re-planning during the inspection.

\end{abstract}


\section{Introduction}

\begin{figure}[!tb]
   \includegraphics[width=\columnwidth]{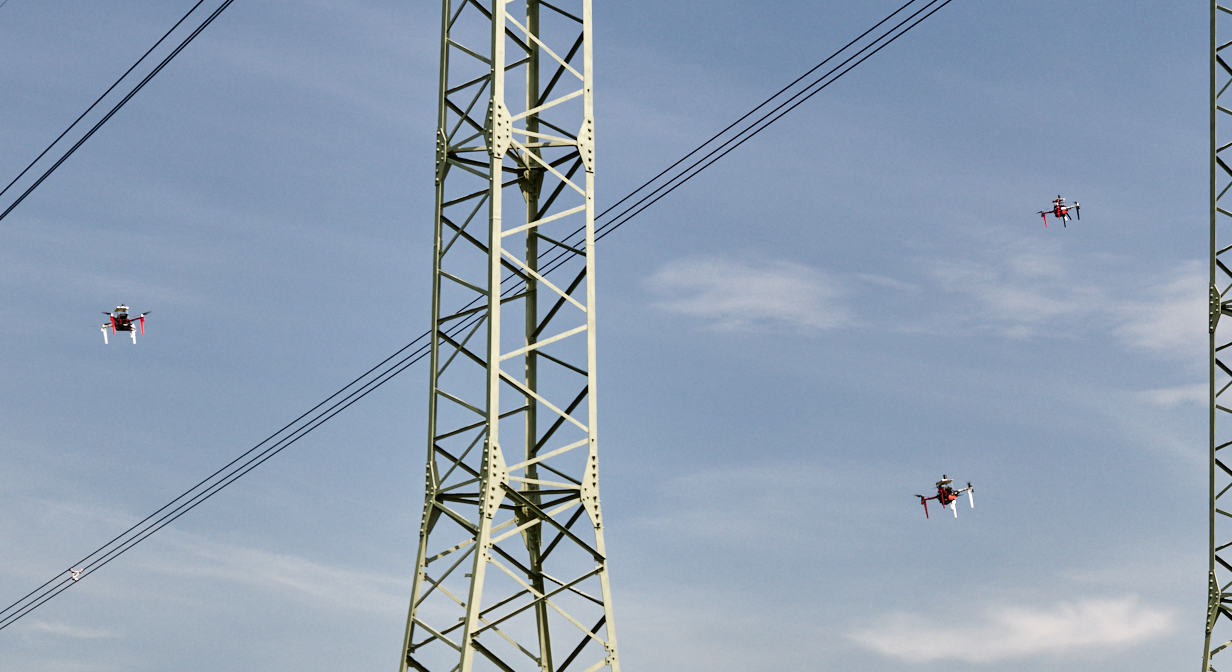}
   \caption{Our MRS research platforms~\cite{hert2022mrs} capable of performing visual inspection on a power transmission line. Source: Neugebauer, CTU in Prague.\label{fig:uav_power_line_inspection}}
   \vspace*{-1em}
\end{figure}

The herein addressed problem is vehicle fault tolerance and robust inspection planning for the \emph{Power Transmission Line} (PTL) inspection, formulated as a generalization of the combinatorial \emph{Traveling Salesman Problem} (TSP)~\cite{nekovar21}.
In further text, the previously formulated generalization shall be referred to as the \emph{Multiple-route Set TSP} (MS-TSP).
When a failure probability of any of the employed \emph{Unmanned Aerial Vehicles} (UAVs), as depicted in Fig.~\ref{fig:uav_power_line_inspection}, is non-negligible, the overall probability of a single-vehicle failure or flight interruption increases with the number of vehicles.
Hence, fault-aware planning is necessary to maximize the success of the inspection mission.
Despite the desirable minimal number of vehicles employed, a limited redundancy should be taken into account in planning.
The most fault-prone part of an electrically powered UAV platform is the battery. 
Measuring the battery's remaining lifetime is difficult as it exhibits non-linear behavior in current and voltage with its life cycle and charge.
Battery fault thus might occur at the beginning of the flight, but it might also happen at any time.
Therefore, our goal is to find a robust inspection plan for multiple multi-copters allowing possible re-planning if a vehicle has to interrupt its mission before finishing the inspection.
The problem addressed is considered a major aspect of the objectives within the EU-funded AERIAL-CORE project.

The transmission line inspection is performed during vehicle flights close to the power lines.
It is advantageous to perform such an inspection flight between segments of the power lines defined by the pylons~\cite{mendenez16vision}.
The problem can be thus defined as the coverage planning problem to visit the transmission line segments, where each segment is a target to be visited.
In the inspection, the segments are organized into a sequence of segments; thus, we formulate the sequencing problem as a variant of the TSP.

A solution of the TSP is a sequence of visits to the given targets that is further generalized to targets grouped into sets, where only a single visit to some target in a set is required, but all sets have to be visited.
Furthermore, inspecting all the sets using a single small UAV with a limited battery budget might not be possible.
Therefore, we generalize the regular TSP formulation to consider the limited maximal flight budget \Tmax{} given by the UAV's battery and consider multiple vehicles to cover all segments with the start and end of the respective tours at the common depot.
As there is a possibility of UAV failure during the inspection flight, it is advantageous to address the planning so that a recovery is possible using other vehicles to finish the inspection goals, as any of the vehicles can fail at any time.
Such a generalized formulation of the TSP is used to describe the addressed PTL inspection planning problem with the vehicle failure tolerance and recovery, where the vehicle is requested to visit each segment of the PTL exactly once but in an arbitrary direction.
The travel cost to the particular start of the segment inspection is also relevant as it is a part of the battery budget \Tmax{}.
\begin{figure}[!htb]
   \vspace{-2em}
   \includegraphics[width=\columnwidth]{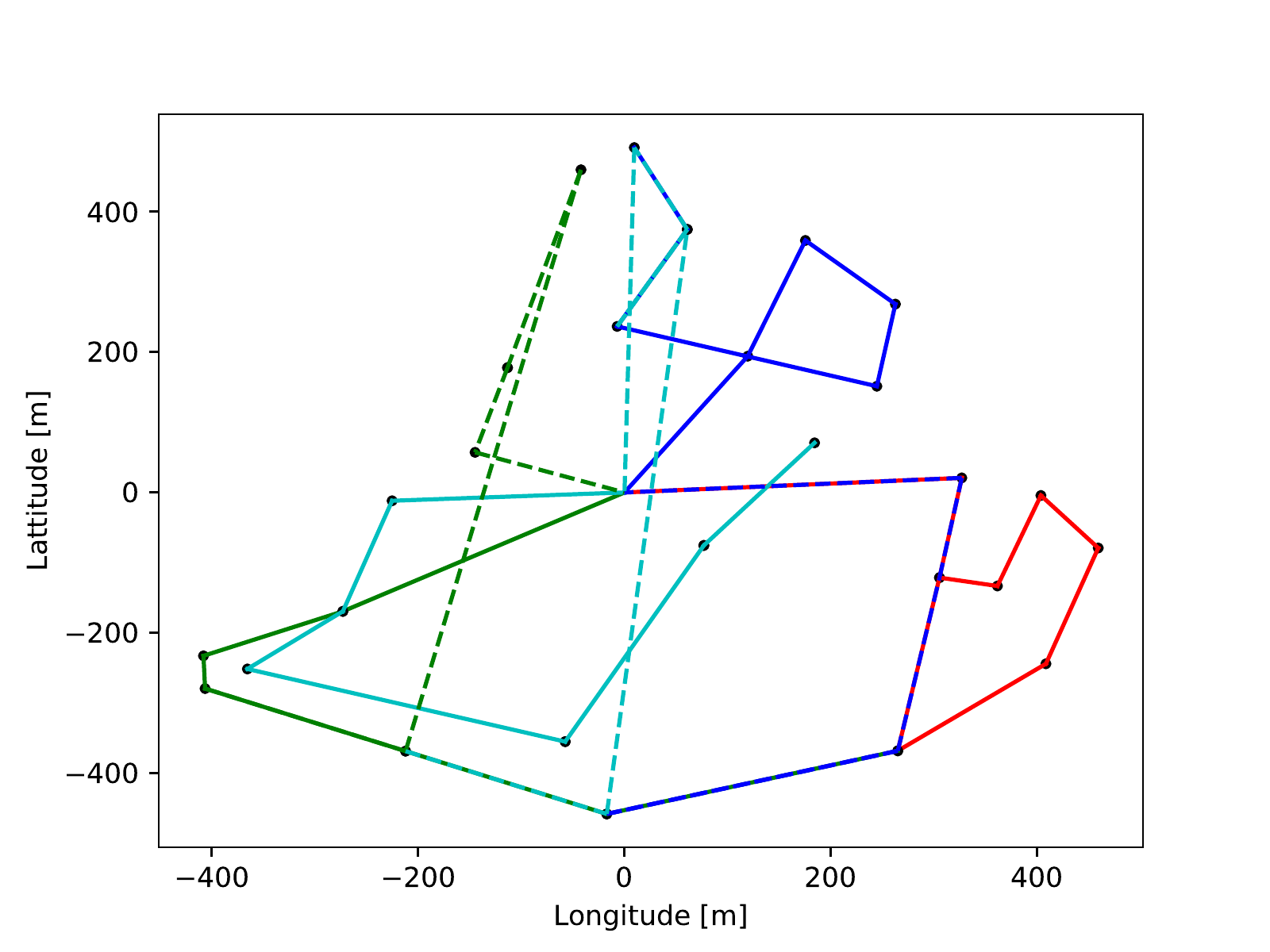}
   \vspace{-2em}
   \caption{An example of the planning and re-planning of an inspection plan around Nechranice power line substation, located at 50°20'44.7"N 13°19'30.9"W, where the vehicle's failure occurred at \SI{50}{\percent} of the maximal route time.} \label{fig:sol_map}
   \vspace{-1em}
\end{figure}
An example of the re-planning is depicted in Fig.~\ref{fig:sol_map}.
The main contribution of the presented short report is considered to be in the evaluation of the three cost functions and robustness of found inspection plans, one of which is novel.
Two of the compared cost functions are standard Minimum Maximum and Minimum Sum, while third one is a weighted sum of the two.
Presented evaluation suggests that our novel cost function provides plans superior in the required criteria.

The remainder of the paper is organized as follows.
A brief overview of related work on power transmission line inspection and existing related formulations of the TSP is provided in the following section.
The addressed robust planning problem is defined in Section~\ref{sec:problem}.
The evaluation methodology of employed combinatorial meta-heuristic for the MS-TSP using different cost functions is discussed in Section~\ref{sec:costs}.
Evaluation results are presented in Section~\ref{sec:results}.
Concluding remarks and discussion of future work are detailed in Section~\ref{sec:conclusion}.

\section{Related Work}\label{sec:related}

Inspection planning problems have been already formulated as variants of the TSP~\cite{galceran13coverage}.
The PTL inspection with the maximum flight distance limit has been addressed using the TSP-based formulation in~\cite{path-alg-uav}.
In our previous work~\cite{nekovar21}, we propose a TSP-based formulation of the power line inspection planning referred to as the \emph{Multiple Set TSP} (MS-TSP).
A multi-vehicle inspection can be formulated as a variant of the \emph{Multiple TSP} (M-TSP)~\cite{bektas06omega} that can be solved using the ILP~\cite{miller60ilptsp}, which is, however, very computationally demanding.
Therefore, a less demanding approach has been proposed by employing a combinatorial metaheuristic called \emph{Greedy Randomized Adaptive Search Procedure} (GRASP)~\cite{guemri16or,kitjacharoenchai2019}.
An overview of M-TSP heuristics is provided by~\cite{cheikhrouhou21mtsp}.
However, our former approach~\cite{nekovar21} differs from the existing variants of the M-TSP in terms of the set generalization and constrained travel budget.

Fault-tolerant UAV control for UAVs has been addressed by several approaches~\cite{tolerant-review} and (re)planning on UAV fault in a formation flight is discussed in~\cite{6315363}, where the authors modify trajectories of formation, so damaged UAVs can follow the reference.
On-board UAV fault detection and diagnostics based on deep learning are proposed in~\cite{9197071}, enabling plan modification and UAV crash prevention in real-time.
In~\cite{5675979}, the authors propose a sliding-mode control for UAVs in the case of external disturbances or actuator fault so that some control performance is achieved on fault or disturbance rejected on fault-free conditions.
The authors of~\cite{DEBENEDETTI201714} discuss the self-organization of UAV flock during data-acquiring terrain monitoring missions using a decentralized solution. 
Fault tolerance on a fault of single or multiple UAVs is achieved by re-planning and re-scouting terrain to avoid data loss due to failure.
In this work-in-progress report, we tackle the fault-tolerant inspection planning based on the MS-TSP formulation.

\section{Problem Statement}\label{sec:problem}
In the addressed PTL inspection, the goal is to traverse all given PTL segments while maximizing the time range for the eventual re-planning as a fault-recovery action.
The segments are defined as the power lines between two pylons, where each segment has to be traversed in a single run to complete its inspection. 
However, the traversing direction can be arbitrary.
The visit to each segment can be formulated as a vertex visit in a set of possible traversing directions, where every set should be visited exactly once, which impels the problem can be formulated as the Generalized TSP (or the Set TSP)~\cite{nekovar21}.

The practical constraints on the UAV behavior for planning are estimated using the \emph{pen-and-paper} model proposed in~\cite{penpaper} for a DJI Mavic 2 platform.
The UAV velocity during flight between the inspection target segment end-points \Vmax{} and velocity during the inspection \Vinsp{} are used in estimating flight times, and a combination of these is then used with the model to estimate the battery budget consumption during flight.

Furthermore, we assume the initial inspection plan is not redundant regarding the number of UAVs, and inspecting all the segments would not be possible using a lower number of vehicles.
As the UAV fault can happen at any time, and the faulty UAV will not be able to complete its inspection plan, the unfinished segments would need to be inspected by the other vehicles.
On vehicle fault, the current target being inspected by the faulty vehicle is assumed to be un-inspected, while remaining vehicles will finish inspection of their current respective targets before continuing using re-planned flight routes.
Our goal is to successfully re-plan in real-time during the flight the inspection plan to visit all the segments using the remaining non-faulty UAVs.
As the number of inspected segments decreases between the beginning and end of each inspection flight, so do the remaining battery budgets on UAVs.
Therefore, we are facing a coupled problem where we aim to maximize the time range where the re-planning might be possible but also minimize the time needed for the inspection of all segments.
However, even at the end of an inspection flight, some collective battery budget is left; the problem is reduced to maximizing the size of a time range from some time point in flight to the end.

The goal of the addressed fault-tolerant planning and the inherent robustness of the inspection plan is to seek an initial inspection plan maximizing the available time range of recovery from a failure of any single UAV and thus finish the inspection of the segments.
We propose to employ our existing GRASP-based solver to initially evaluate several cost functions to achieve the desired challenging solution of the fault-tolerant planning.
In theory, a suitable cost function should exist, which by taking advantage of the inherent randomness of GRASP, increases the chances of finding such a plan.

\section{Cost functions}\label{sec:costs}

Three cost functions were chosen for the initial evaluation of the proposed approach with the following notation.
$T$ denotes a tour that is a part of a set of all tours $\mathcal{T}$.
The unconstrained battery cost of the tour $T$ is refered to as a function $c(T)$.
The three costs are as follows.
\begin{enumerate}
   \item The \emph{Minimum-Maximum} (MinMax) cost~\eqref{eq:minmax} denoted $c_{minmax}$ function minimizes the maximum battery budget cost of the most battery consuming tour.
\begin{equation}\label{eq:minmax}
    c_{minmax} =\text{max}(c(T))
\end{equation}
\item The \emph{contrained Minimum-Sum} (c-MinSum) cost~\eqref{eq:minsum} function $c_{minsum}$ adds a soft constraint on the maximal battery budget cost to the battery budget cost, which enables us to minimize the total sum of the consumed battery over all the tours while adhering to the battery budget constraint.
The soft constraint is violated if the tour cost $c(T)$ exceeds the maximum allowed cost \Tmax, as indicated in~\eqref{eq:soft}. 
The c-MinSum cost $c_{minsum}$ constraints the cost of a single tour $T$ as $c_{con}(T)$,
\begin{equation}\label{eq:minsum}
    c_{minsum} = \sum_{T \in \mathcal{T}} c_{con}(T)
\end{equation}
\begin{equation}\label{eq:soft}
    c_{con}(T) = 
    \begin{cases}
      c(T) & \text{if } c(T) \leq \Tmax,\\
      c(T) + (c(T)-\Tmax)k_c & \text{if } c(T) > \Tmax,
    \end{cases}
\end{equation}
where $k_c = 10^3$ is chosen to be a sufficiently large number with respect to the tour costs.
\item A combined cost~\eqref{eq:mincomb} of the c-MinSum cost and the weighted MinMax cost divided by the number of tours $n_t$.
\begin{equation}
   \label{eq:mincomb}
   c_{mincomb} = c_{minsum} + \frac{1}{n_t}c_{minmax}
\end{equation}
\end{enumerate}
The reasoning behind the combined cost is that while the constrained c-MinSum cost function provides the lowest combined battery consumption, and thus provides the highest left-over budget at the end of the inspection, it performs poorly at balancing the consumption between UAVs.
The MinMax cost function on the other hand does the opposite, it balances the costs to minimize the maximal tour cost, which is desirable, as any UAV can be interrupted.
The third cost function is a weighted sum of the original cost functions.
By favorizing the c-MinSum cost part, while adding the balancing factor of MinMax, the combined cost function should provide a trade-off between both of them.

\section{Results}\label{sec:results}

The evaluation of the proposed cost functions has been performed on the real-world scenario~\cite{mrs-mstsp} of the PTLs originating from the substation Nechranice, courtesy of ČEPS, a.s. (Czech power transmission infrastructure institution, a member of the AERIAL-CORE advisory board).
The battery budget of a single UAV is given by \Tmax{}.
Flight between inspection segments is limited by the maximum flight velocity $\Vmax = \SI{5}{\meter\per\second}$ and the inspection flight is limited by the maximum inspection velocity $\Vinsp = \SI{1}{\meter\per\second}$.
Additional flight dynamics are considered negligible and not used in the battery estimation model because the flight distances are in order of tens and hundredths of meters.
For the pen-and-paper model~\cite{penpaper}, battery budget \SI{100}{\percent}, the maximum travel distance \Dmax{} we can plan for four UAVs is \SI{700}{\meter}.
Hence, for the initial computational evaluation using the real-world scenario, the battery budget is relaxed.
The number of UAVs $n_t = 4$ is fixed, and the battery budget varies so that all the cost functions can provide a valid initial solution. 
\begin{figure*}
\centering
{
\setlength{\tabcolsep}{0.1pt}
\begin{tabular}{ccc}
\includegraphics[width=0.32\textwidth]{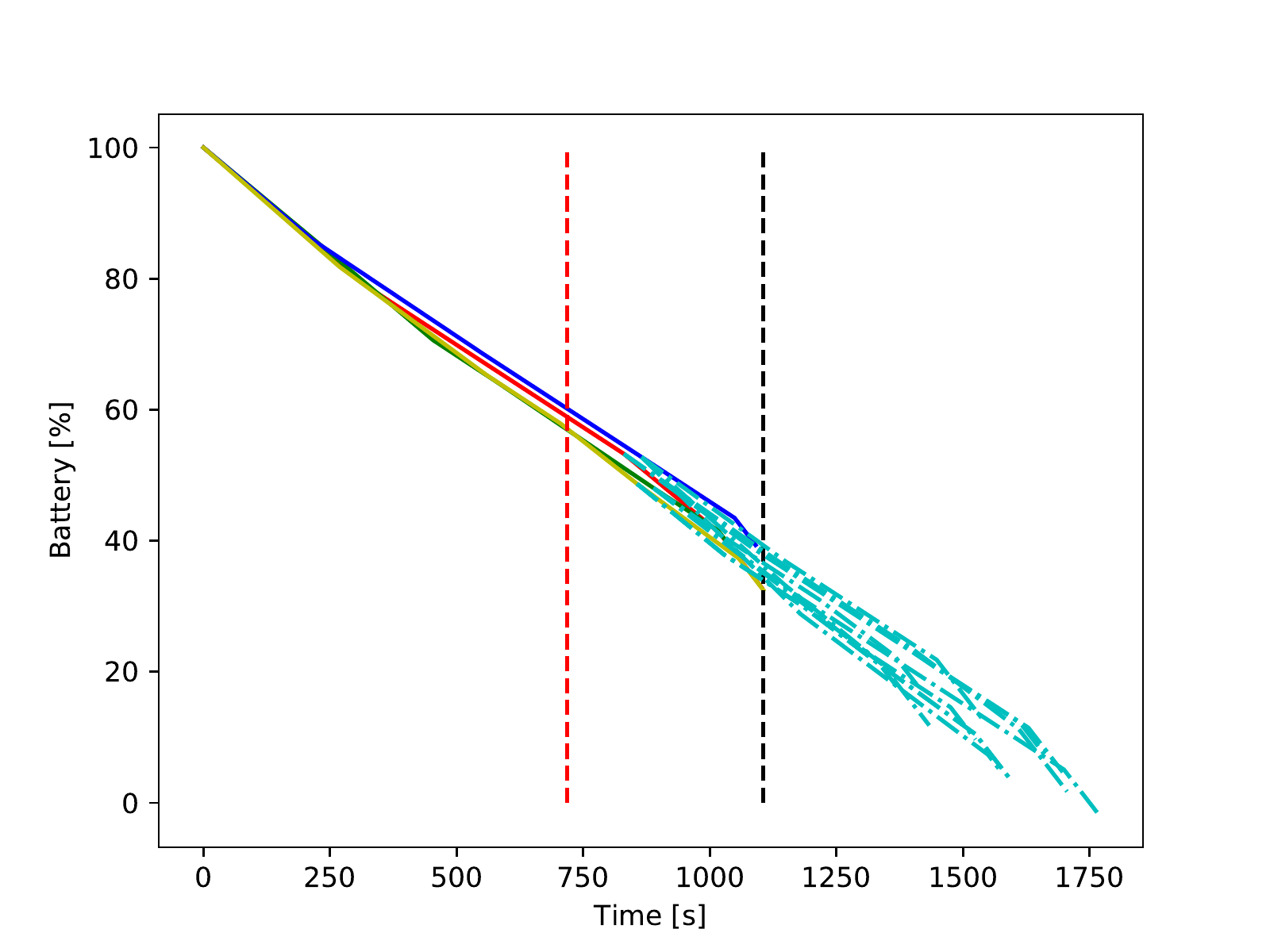} &
\includegraphics[width=0.32\textwidth]{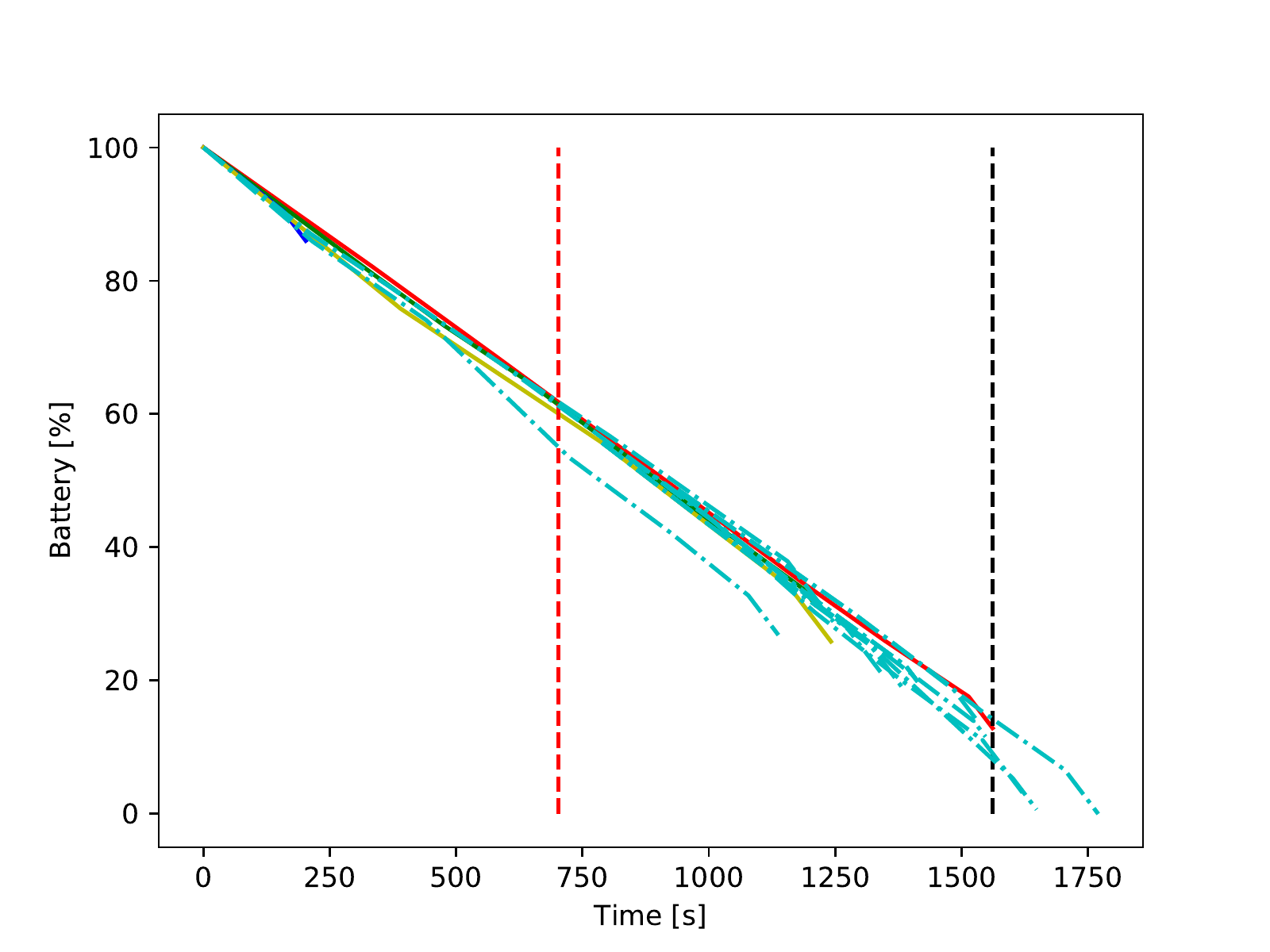} &
\includegraphics[width=0.32\textwidth]{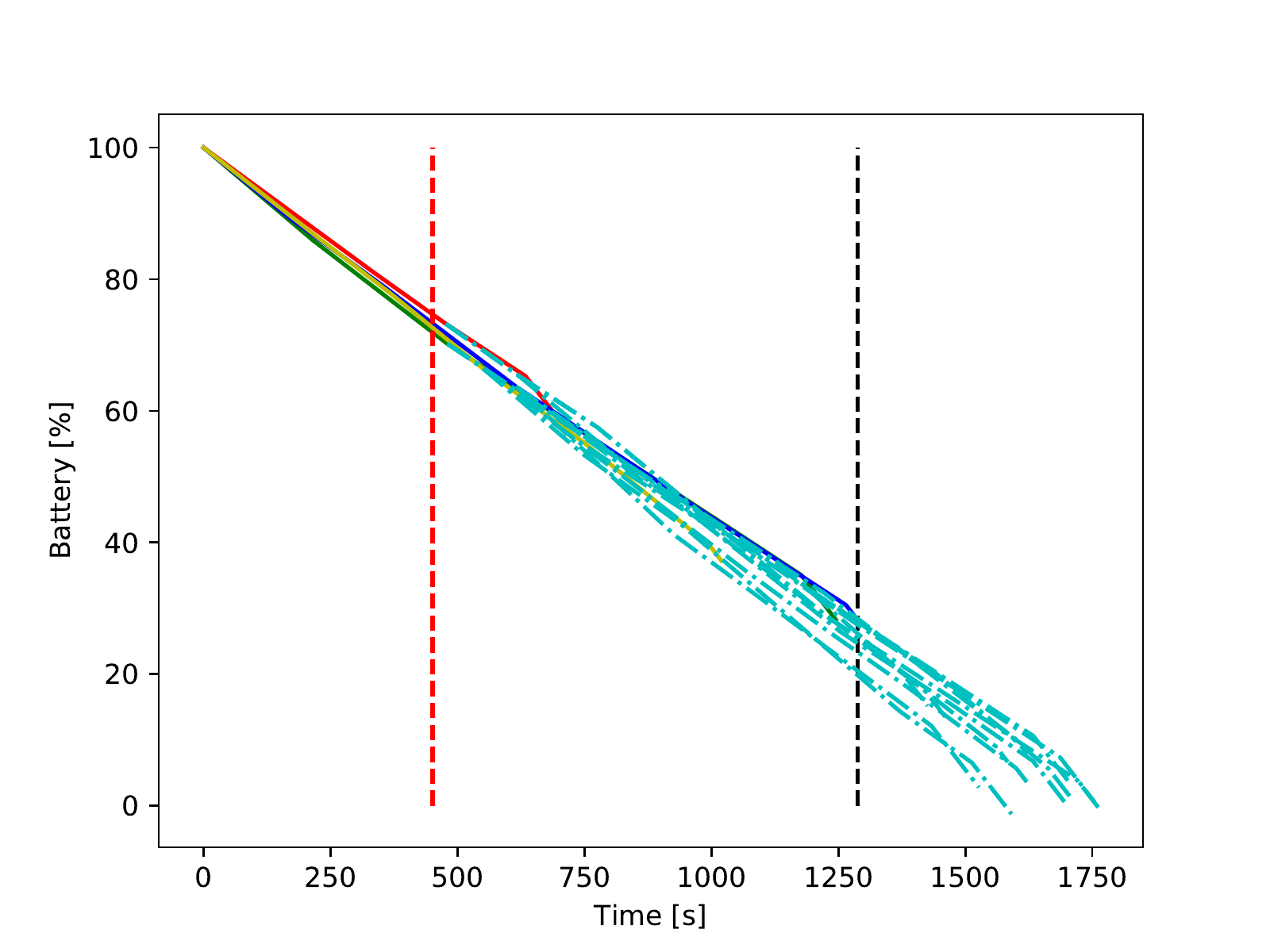} \\
\end{tabular}
}
\caption{\small \label{fig:replans} 
Battery consumption in time for $\Dmax = 500$ obtained using the MinMax, c-MinSum and combined cost functions. The re-plan windows are, respectively: \SI{20}{\percent}, \SI{58}{\percent} and \SI{66}{\percent}. Red vertical line indicates earliest time re-planning is possible, black vertical line indicates maximal route time. Battery budget consumptions of plans are solid color lines, of re-plans dashed cyan lines.}
\end{figure*}
Examples of the found plans and subsequent re-plans are shown in~\cref{fig:replans}, where the vertical red line denotes the minimal time from the inspection start at which any of the UAVs can fail and re-planning is possible using the remaining UAVs.
The battery budget consumptions of the plans are shown as the solid color lines for all the UAVs, while the subsequent re-plan consumptions are shown as the cyan dotted lines for clarity.
Computational results are depicted in \cref{tbl:results}.
The value $n_{seg}$ indicates number of inspection targets in the area.
Best re-planning window size for each evaluation scenario is indicated in bold.
It is the percentage of maximal tour time in plan starting from the end, and seen in \cref{fig:replans}

\begin{table}[!htb]\centering
   \caption{Computational Results}\label{tbl:results}
   \vspace{-1em}
   \scalebox{0.45}{
      \resizebox{1\textwidth}{!}{
	 \begin{tabular}{rrccccc}
	 \noalign{\hrule height 1.1pt}\noalign{\smallskip}
	    \multicolumn{1}{c}{\Dmax} 
	    & \multicolumn{1}{c}{$n_{seg}$}
	    &\multicolumn{1}{c}{\Cmax} 
	    & \multicolumn{1}{c}{$n_{t}$}
	    &\multicolumn{1}{c}{MinMax} 
	    &\multicolumn{1}{c}{c-MinSum}
	    &\multicolumn{1}{c}{MinComb} \\
	    \multicolumn{1}{c}{[\si{m}]} && \multicolumn{1}{c}{[\si{\percent}]}& 
	    & \multicolumn{1}{c}{[\si{\percent}]}& \multicolumn{1}{c}{[\si{\percent}]}& \multicolumn{1}{c}{[\si{\percent}]}
	    \\
	 \noalign{\smallskip}\hline\noalign{\smallskip}
    {500}&{15}&{1\,00}&{4}&{20}&{58}&\textbf{66}\\ 
    {700}&{28}&{2\,00}&{4}&{17}&{86}&\textbf{96}\\ 
    {1\,000}&{43}&{2\,50}&{4}&{14}&{63}&\textbf{65}\\ 
    {1\,200}&{51}&{3\,00}&{4}&{23}&{56}&\textbf{59}\\ 
    {2\,000}&{76}&{5\,00}&{4}&{28}&{51}&\textbf{70}\\ 
	 \noalign{\smallskip}\noalign{\hrule height 1.1pt}
	 \end{tabular}
      }
   }
\end{table}

The proposed approach has been implemented in C++ and executed on the Ryzen 7 4750U CPU with \SI{32}{\giga\byte} of RAM.
The planning and re-planning times are in order of seconds for all the instances.
The GRASP MS-TSP solver was run 50 times for the initial planning and re-plannings.
The whole planning and re-planning procedure have been repeated ten times for each instance, and the best result is presented.

\begin{figure}[!tb]
   \includegraphics[width=\columnwidth]{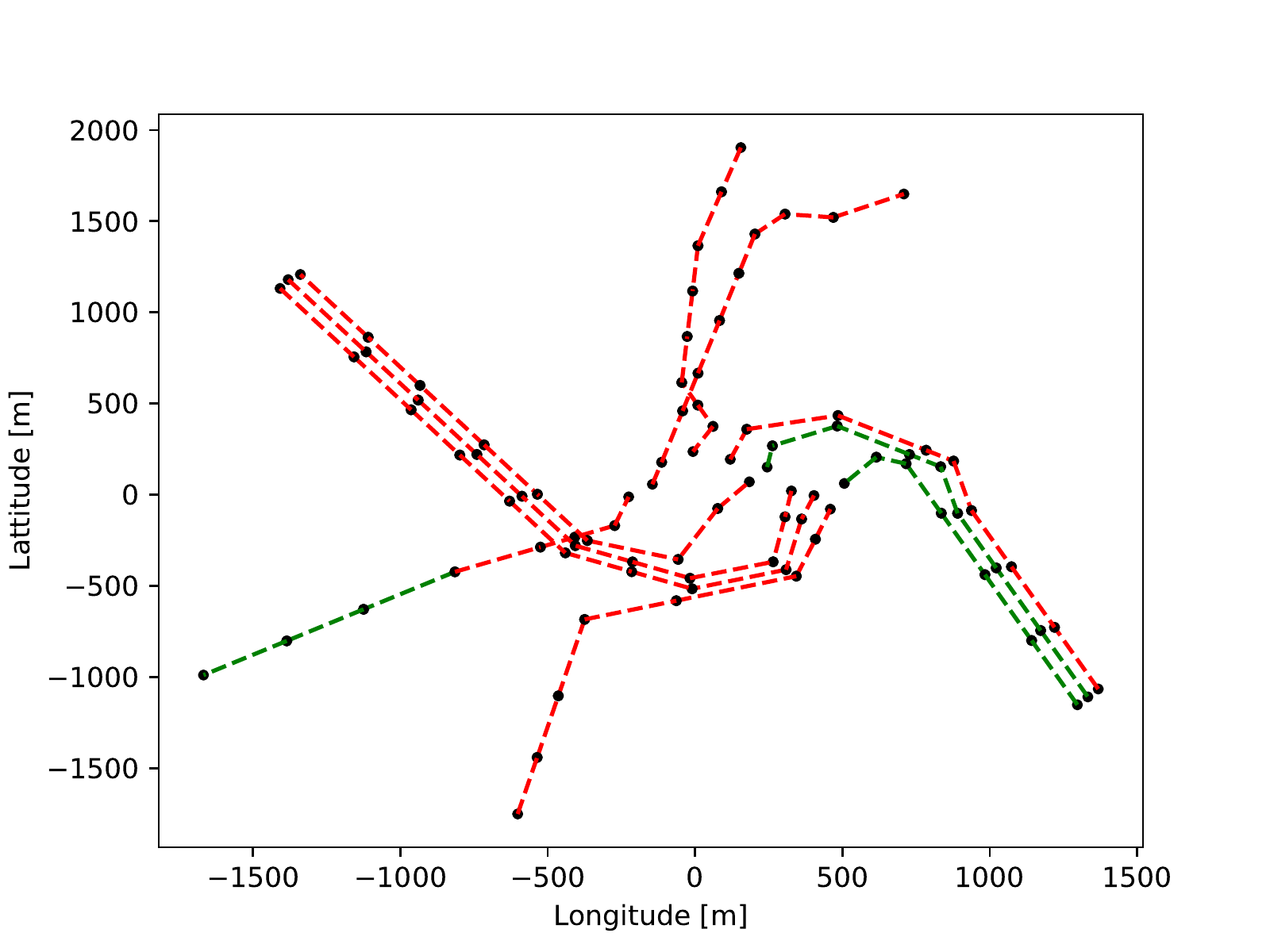}
  \vspace*{-2em}
   \caption{An example of the planning and re-planning at $\Dmax=2\,000$ using combined cost function. Vehicle interruption occurred at \SI{70}{\percent} of the maximal route time.} \label{fig:sol_big}
  \vspace*{-1em}
\end{figure}

The combined cost function performs best at maximizing the re-planning time window.
In~\cref{fig:sol_big}, we can see target inspection segments covered by original plan as green lines and by re-plan as red lines.
The MinSum cost function is better suited for our purpose than the MinMax cost function.
The combined cost function improves the performance and supports our initial assumption about the cost function influence, as we can see in~\cref{tbl:results}.

\section{Conclusion}\label{sec:conclusion}
The performance of the GRASP-based fault-tolerant PTL inspection planning problem was proposed and evaluated in this work-in-progress paper. The main benefit of the proposed combined cost function is the maximization of the available time window where we can successfully recover from any single UAV failure during the inspection flight.
While not considering other qualities of the initial plan, such as the ratio of the elapsed time and consumed battery budget or UAV positions during the flight, we show that the robustness can be influenced by using a suitable cost function.
We also suggest that the inherent randomness of the GRASP-based solver can provide, over multiple re-runs, a plan superior in the criteria to the one that would be truly optimal with respect to the cost function, obtainable by formulating the problem using Integer Linear Programming (ILP).
Further research is directed toward taking advantage of the aforementioned plan qualities and formulating a suitable solver.

\bibliographystyle{IEEEtran}
\bibliography{main}
\end{document}